\documentclass{INTERSPEECH2023}

\interspeechcameraready

\title{Relationship between auditory and semantic entrainment using Deep Neural Networks (DNN)}
\name{Jay Kejriwal$^{1,2}$, Štefan Beňuš$^{1,3}$}
\address{
  $^1${Institute of Informatics}, {Slovak Academy of Sciences}, {Slovakia} \\
$^2${Faculty of Informatics and Information Technology}, {Slovak Technical University}, {Slovakia}\\
  $^3${Constantine the Philosopher University}, {Nitra, Slovakia} }
\email{jay.kejriwal@savba.sk, sbenus@ukf.sk}

\begin{document}

\maketitle
 
\begin{abstract}
The tendency of people to engage in similar, matching, or synchronized behaviour when interacting is known as entrainment. Many studies examined linguistic (syntactic and lexical structures) and paralinguistic (pitch, intensity) entrainment, but less attention was given to finding the relationship between them. In this study, we utilized state-of-the-art DNN embeddings such as BERT and TRIpLet Loss network (TRILL) vectors to extract features for measuring semantic and auditory similarities of turns within dialogues in two comparable spoken corpora of two different languages. We found people's tendency to entrain on semantic features more when compared to auditory features. Additionally, we found that entrainment in semantic and auditory linguistic features are positively correlated. The findings of this study might assist in implementing  the mechanism of entrainment in human-machine interaction (HMI).
\end{abstract}

\section{Introduction} 
Entrainment is the tendency of a speaker to adjust some properties of a speaker’s features to match the interlocutor’s characteristics. It has been found to correlate with positive social attributes such as likeability \cite{ireland2011a}, task success \cite{reitter2007a}, and even rapport with a robot \cite{lubold2015a}. According to the psycholinguistics literature, entrainment affects various linguistic dimensions, such as lexical choice \cite{brennan1996a}, syntactic structure \cite{reitter2006a}, or acoustic-prosodic features \cite{levitan2011b}. 

Several studies have investigated the effects of entrainment utilizing different modalities and implemented it in Spoken Dialogue Systems (SDS) \cite{lubold2015a,levitan2016a,levitan2021a}. In SDS, speech entrainment functionality would enable machines to dynamically entrain and disentrain on various auditory features, which might result in more efficient, successful, natural, and pleasing interactions. Similarly, implementing semantic entrainment functionality would enable machines to align semantically with humans resulting in more meaningful conversations. An essential first step toward effectively implementing entrainment in SDS is understanding how entrainment works at different linguistic levels and what their relationships are. Understanding these variations will allow us to weigh them meaningfully when they are combined to develop SDS systems equipped with effective entrainment functionalities. 

Entrainment has previously been studied independently using linguistic-related parameters \cite{ward2007a} or paralinguistic-related parameters \cite{levitan2011a,levitan2015a}. Additionally, researchers have started exploring the correlation between entrainment at different linguistic levels. For instance, \cite{patel2022a} explored the relationship between prosodic, lexical, semantic, and syntactic entrainment among individuals with autism spectrum disorder (ASD). The results revealed distinct patterns of prosodic and lexical entrainment. Similarly, \cite{ostrand2021a} explored the correlation between acoustic-prosodic and syntactic entrainment within a dialogue. They reported speakers entrain on some but not all features within a linguistic level. Furthermore, \cite{rahimi2017a} reported correlations between acoustic-prosodic and lexical entrainment in group conversations. On the contrary, \cite{weise2018a} found that none of the acoustic-prosodic and lexical entrainment measures were meaningfully correlated, clustered, or exhibited principal components. Hence, the results of studies exploring the relationship of entrainment at different levels are inconclusive.
 
In a recent study \cite{liu2021a}, DNN embeddings were used to explore the relationship between acoustic-prosodic and semantic entrainment. The authors proposed measures of “semantic similarity” of dialogues using BERT embeddings trained on a Chinese spoken corpus. They reported an inverse relationship between them: interlocutors did not adjust prosodic features when their semantics were closer to their partners. However, these results and their wider impact on SDS applications should be interpreted with caution since there were three limitations to the given study. First, the question-response system in Chinese conversations was analyzed. The authors did not provide a cross-linguistic comparison, which would allow observing the trends and underlying patterns by comparing auditory and semantic entrainment in different languages. Second, the authors introduce \textit{convergence} and \textit{synchrony} as entrainment metrics. \textit{Convergence} implies people become more similar over the period of time. \textit{Synchrony} means people are consistently behaving in similar way. The authors did not consider proximity as an entrainment metric which is helpful in understanding if two people are getting semantically closer to each other at a given time. In a session that displays proximity, the speaker turns are more similar to the immediately adjacent turns of the interlocutor than to other random interlocutor’s turns \cite{levitan2015a}. Information about proximity might be valuable for turn-to-turn implementation of entrainment into automatic SDS. Last, the authors did not report if BERT embeddings were normalized or not. Usually, BERT embeddings are not normalized and utilizing Pearson's correlation can provide inconsistent results. There is a high degree of sensitivity in Pearson's \textit{r} to even minor deviations from normality, where an outlier can hide an underlying association \cite{zhelezniak2019a}. Using a novel approach in this study, we describe linguistic information and analyze the entrainment relationship between two different linguistic levels by utilizing different entrainment metrics (proximity, convergence, and synchrony) on two different spoken corpora using different languages. 

Empirical studies exploring the entrainment relationship between different linguistic levels have found variable results so far. There might be three possible reasons associated with it. First, entrainment in linguistic levels has been analyzed using different methods. For example, in \cite{ostrand2021a}, the authors measured acoustic-prosodic entrainment using the metrics proposed in \cite{levitan2011b}, which measures correlations among adjacent turns. In contrast, they analyzed syntactic entrainment with generalized logit mixed-effect models (GLMM) \cite{glmm}. Second, different toolkits are utilized for feature extraction.  For example, in \cite{ostrand2021a}, researchers used the PRAAT toolkit \cite{boersma2009a} for extracting 323  temporal and acoustic-prosodic features, whereas \cite{patel2022a} derived pitch, intensity, and rhythm-related features using the contour-based, parametric, and super positional intonation stylization (CoPaSul) toolkit that uses some different feature extraction and manipulation approaches \cite{reichel2020a}.  Lastly, researchers measured similarity using different units of analysis. In \cite{weise2018a}, authors measure acoustic-prosodic entrainment on the inter-pausal unit (IPU)\footnote{IPU is a pause-free unit in turn separated by at least 50 ms. of silence}, whereas they measured lexical entrainment using \textit{n}-gram sequences.  In this study, we will extract features and measure entrainment using the same methodology in an effort to limit the mentioned sources of variability in results.

Finally, empirical findings on entrainment suggest it is a complex phenomenon where people entrain/dis-entrain on different para-linguistic features \cite{levitan2011b}. Earlier studies on entrainment have utilized paralinguistic features that incorporate spectral, temporal, and acoustic-prosodic features. A DNN embedding can solve the problem of fragmentation in para-linguistic features. DNN embedding is a method used to represent discrete variables as continuous vectors. DNN embedding using textual modality such as transformer \cite{NIPS2017_3f5ee243} is immensely popular and has broader applications in NLP applications. Similarly, DNN embedding using auditory modality has provided promising outcomes in improving the performance of automatic speech recognition and other applications. In \cite{shor2020a}, the TRILL vector was proposed, which creates embeddings based on a CNN architecture that uses triplet-loss representation. This approach maps audio segments that appear nearer in time to be nearer in the embedding space. A comparison of different auditory features such as Low-level descriptors (LLD) features, spectral features, and DNN audio embeddings (x-vectors, TRILL vectors) was presented in \cite{kejriwal2022a}. In comparison to different auditory features, TRILL vectors provided greater classification accuracy in this study. Hence, we employ this method in our work to compare the acoustic and semantic entrainment. 

In sum, research into speech entrainment has so far been fragmented, with numerous individual features and measures of similarity being used, but no attempts have been made prior to our knowledge that measures auditory similarity using DNN embeddings. With a long-term goal to develop an effective SDS, we analyze in this study auditory and semantic entrainment in comparable corpora of conversational speech in English and Slovak. Our paper makes three main contributions. First, we measured entrainment in conversational corpora using state-of-the-art DNN embeddings on semantic and auditory levels. Second, we explore the relationship between the two levels using the same methodology. Finally, the experimental result shows that entrainment in both levels is correlated positively in both spoken corpora.

\section{Data and features} \label{Data and features}
In this section, we describe two task-oriented spoken language corpora we analysed in the current study, how we extracted semantic and auditory features from them, and how we calculated metrics for measuring auditory and semantic entrainment.
\subsection{Dataset} 

\subsubsection{Columbia Games Corpus}
The Columbia Games Corpus \cite{hirschberg2021a} consists of 12 spontaneous dyadic conversations between native Standard American English (SAE) speakers. Participants included thirteen individuals (six females and seven males); eleven participated in two sessions on different days and with other partners. Each dyad played four computer games of two kinds: Cards games and Objects games involving communication and teamwork. The subjects did not have visual contact due to a curtain placed between them ensuring verbal communication only. Twelve sessions were recorded, totaling 9 hours and 13 minutes. The subset of the Columbia Games Corpus most closely resembling spontaneous task-related conversations, namely the Objects game, was used for the current study, which roughly comprises 4.3 hours of speech data.

\subsubsection{SK-Games Corpus}

The SK-games corpus \cite{be2012a} is identical to the Objects games of the Columbia Games Corpus for SAE, except for changes in some screen images and their locations. The corpus contains nine dyadic conversations recorded by native speakers of Slovak. Eleven speakers (five females and six males) participated in the study; seven participated in two sessions, each with a different partner. The corpus involves 6.3 hours of spoken dialogue.

\subsection{Feature extraction}

The semantic and auditory linguistic levels of entrainment are analysed in each corpus. To extract semantic features, each turn in the dialog is encoded into a fixed-length vector (embeddings). For the Columbia games corpus, we used a neural network-trained model (SBERT) \cite{reimers2019a}, representing 768 one-dimensional semantic features for each turn. Similarly, for the SK-Games corpus, we used the Slovak masked language model called SlovakBERT \cite{pikuliak2021a} where each turn is encoded into 768 one-dimensional semantic features. Furthermore, to extract auditory features for each turn, the TRILL vector \cite{hoffer2018a} is used, representing 512 one-dimensional auditory features per turn. Since the TRILL vector model is language-independent, we used the same model on both the spoken corpora.

\subsection{Entrainment metrics}

In \cite{levitan2011a}, the authors introduced three measures of entrainment:  \textit{Proximity} describes the similarity of interlocutor’s speech at turn exchanges.  \textit{Convergence} quantifies the tendency when two speaker’s speech becomes more similar throughout the conversation.  \textit{Synchrony} describes the entrainment by direction where speaker’s prosodic features become correlated to his/her interlocutor. Based on the definition of the given metrics we used the same metrics for the current study. In earlier studies, absolute values were used to measure entrainment on acoustic-prosodic features. Since we are using DNN embeddings in the current study, the metrics are re-defined.

\textit{Proximity} is measured using paired t-tests on two sets of differences: a set of adjacent distance (Eq.\ref{equation:eq1}) and another corresponding set of non-adjacent distance (Eq.\ref{equation:eq2}). Adjacent distance is the cosine distance between speaker's embeddings and his/her conversational partners adjacent embeddings. On the other hand, non-adjacent distance is the cosine distance between the embeddings of a speaker and other random non-adjacent embeddings of his/her conversational partner. For ten random turns of another speaker, we measured the non-adjacent distance and calculated the mean. If the cosine distance of the adjacent distance is greater than the non-adjacent distance, we can infer that speakers are getting closer to each other.  
 \begin{align}
adjacent\ distance=\cos (A,B) = \frac{A \cdot B}{|A||B|}
\label{equation:eq1}
\end{align}

\begin{align}
non-adjacent\ distance=\ \sum_{i=1}^{10}\cos({A,B_{rand}})
\label{equation:eq2}
\end{align}

 \noindent \textit{Convergence} is measured by Pearson's correlation between cosine distance between adjacent turns and turn number (time). \textit{Synchrony} is measured using Pearson's correlation on two sets of self-distance of speaker A and B. Self-distance (Eq.\ref{equation:eq3}) of a speaker is measured using cosine similarity between two consecutive turns of the same speaker. 

\begin{align}
self\ distance=\cos({A_i},{A_{i+1}})
\label{equation:eq3}
\end{align}

\section{Results} 
\subsection{Auditory and semantic entrainment using DNN}
\begin{table}[h!]
    \centering
    \resizebox{8cm}{!}{%
\begin{tabular}{ccccccccccccc|}
\cline{2-13}
\multicolumn{1}{c|}{}                  & \multicolumn{4}{c|}{\textbf{Proximity}}                                                                                                                     & \multicolumn{4}{c|}{\textbf{Convergece}}                                                                                                                    & \multicolumn{4}{c|}{\textbf{Synchrony}}                                                                                                \\ \cline{2-13} 
\multicolumn{1}{c|}{}                  & \multicolumn{2}{c|}{Auditory}                                                & \multicolumn{2}{c|}{Semantic}                                                & \multicolumn{2}{c|}{Auditory}                                                & \multicolumn{2}{c|}{Semantic}                                                & \multicolumn{2}{c|}{Auditory}                                                & \multicolumn{2}{c|}{Semantic}                           \\ \hline
\multicolumn{1}{|c|}{\textit{Session}} & \multicolumn{1}{c|}{\textit{t}} & \multicolumn{1}{c|}{\textit{Sig.}} & \multicolumn{1}{c|}{\textit{t}} & \multicolumn{1}{c|}{\textit{Sig.}} & \multicolumn{1}{c|}{\textit{r}} & \multicolumn{1}{c|}{\textit{Sig.}} & \multicolumn{1}{c|}{\textit{r}} & \multicolumn{1}{c|}{\textit{Sig.}} & \multicolumn{1}{c|}{\textit{r}} & \multicolumn{1}{c|}{\textit{Sig.}} & \multicolumn{1}{c|}{\textit{r}} & \textit{Sig.} \\ \hline
\multicolumn{1}{|c|}{1}                & \multicolumn{1}{c|}{4.04}       & \multicolumn{1}{c|}{\textbf{*}}            & \multicolumn{1}{c|}{3.28}       & \multicolumn{1}{c|}{\textbf{*}}            & \multicolumn{1}{c|}{0.02}       & \multicolumn{1}{c|}{}                      & \multicolumn{1}{c|}{0.03}       & \multicolumn{1}{c|}{}                      & \multicolumn{1}{c|}{0.03}       & \multicolumn{1}{c|}{}                      & \multicolumn{1}{c|}{0.03}      &                       \\
\multicolumn{1}{|c|}{2}                & \multicolumn{1}{c|}{-0.05}      & \multicolumn{1}{c|}{}                      & \multicolumn{1}{c|}{4.44}       & \multicolumn{1}{c|}{\textbf{*}}            & \multicolumn{1}{c|}{0.11}       & \multicolumn{1}{c|}{\textbf{+}}            & \multicolumn{1}{c|}{-0.02}     & \multicolumn{1}{c|}{}                      & \multicolumn{1}{c|}{-0.02}     & \multicolumn{1}{c|}{}                      & \multicolumn{1}{c|}{0.08}      &                       \\
\multicolumn{1}{|c|}{3}                & \multicolumn{1}{c|}{-0.28}      & \multicolumn{1}{c|}{}                      & \multicolumn{1}{c|}{1.05}       & \multicolumn{1}{c|}{\textbf{}}             & \multicolumn{1}{c|}{-0.09}      & \multicolumn{1}{c|}{}                      & \multicolumn{1}{c|}{-0.13}      & \multicolumn{1}{c|}{\textbf{+}}            & \multicolumn{1}{c|}{-0.13}      & \multicolumn{1}{c|}{\textbf{+}}            & \multicolumn{1}{c|}{-0.07}     &                       \\
\multicolumn{1}{|c|}{4}                & \multicolumn{1}{c|}{3.05}       & \multicolumn{1}{c|}{\textbf{*}}            & \multicolumn{1}{c|}{5.03}       & \multicolumn{1}{c|}{\textbf{*}}            & \multicolumn{1}{c|}{-0.08}      & \multicolumn{1}{c|}{}                      & \multicolumn{1}{c|}{-0.08}      & \multicolumn{1}{c|}{}                      & \multicolumn{1}{c|}{-0.08}      & \multicolumn{1}{c|}{}                      & \multicolumn{1}{c|}{0.01}      &                       \\
\multicolumn{1}{|c|}{5}                & \multicolumn{1}{c|}{3.01}       & \multicolumn{1}{c|}{\textbf{*}}            & \multicolumn{1}{c|}{3.01}       & \multicolumn{1}{c|}{\textbf{*}}            & \multicolumn{1}{c|}{-0.13}      & \multicolumn{1}{c|}{\textbf{+}}            & \multicolumn{1}{c|}{-0.01}      & \multicolumn{1}{c|}{}                      & \multicolumn{1}{c|}{-0.01}      & \multicolumn{1}{c|}{}                      & \multicolumn{1}{c|}{-0.06}     &                       \\
\multicolumn{1}{|c|}{6}                & \multicolumn{1}{c|}{2.36}       & \multicolumn{1}{c|}{\textbf{+}}            & \multicolumn{1}{c|}{4.28}       & \multicolumn{1}{c|}{\textbf{*}}            & \multicolumn{1}{c|}{-0.02}      & \multicolumn{1}{c|}{}                      & \multicolumn{1}{c|}{-0.04}      & \multicolumn{1}{c|}{}                      & \multicolumn{1}{c|}{-0.04}      & \multicolumn{1}{c|}{}                      & \multicolumn{1}{c|}{0.16}      & \textbf{+}            \\
\multicolumn{1}{|c|}{7}                & \multicolumn{1}{c|}{1.32}       & \multicolumn{1}{c|}{}                      & \multicolumn{1}{c|}{2.87}       & \multicolumn{1}{c|}{\textbf{+}}            & \multicolumn{1}{c|}{0.03}       & \multicolumn{1}{c|}{}                      & \multicolumn{1}{c|}{0.09}       & \multicolumn{1}{c|}{\textbf{+}}            & \multicolumn{1}{c|}{0.09}       & \multicolumn{1}{c|}{\textbf{+}}            & \multicolumn{1}{c|}{0.04}      &                       \\
\multicolumn{1}{|c|}{8}                & \multicolumn{1}{c|}{-0.02}      & \multicolumn{1}{c|}{}                      & \multicolumn{1}{c|}{3.13}       & \multicolumn{1}{c|}{\textbf{*}}            & \multicolumn{1}{c|}{-0.06}      & \multicolumn{1}{c|}{}                      & \multicolumn{1}{c|}{0.05}       & \multicolumn{1}{c|}{}                      & \multicolumn{1}{c|}{0.05}       & \multicolumn{1}{c|}{}                      & \multicolumn{1}{c|}{0.01}      &                       \\
\multicolumn{1}{|c|}{9}                & \multicolumn{1}{c|}{-1.92}      & \multicolumn{1}{c|}{}                      & \multicolumn{1}{c|}{0.15}       & \multicolumn{1}{c|}{}                      & \multicolumn{1}{c|}{0.10}       & \multicolumn{1}{c|}{}                      & \multicolumn{1}{c|}{-0.01}      & \multicolumn{1}{c|}{}                      & \multicolumn{1}{c|}{-0.01}      & \multicolumn{1}{c|}{}                      & \multicolumn{1}{c|}{0.06}      &                       \\
\multicolumn{1}{|c|}{10}               & \multicolumn{1}{c|}{-0.11}      & \multicolumn{1}{c|}{}                      & \multicolumn{1}{c|}{2.27}       & \multicolumn{1}{c|}{\textbf{+}}            & \multicolumn{1}{c|}{0.06}       & \multicolumn{1}{c|}{}                      & \multicolumn{1}{c|}{0.10}       & \multicolumn{1}{c|}{\textbf{*}}            & \multicolumn{1}{c|}{0.10}       & \multicolumn{1}{c|}{\textbf{*}}            & \multicolumn{1}{c|}{-0.06}     &                       \\
\multicolumn{1}{|c|}{11}               & \multicolumn{1}{c|}{-2.19}      & \multicolumn{1}{c|}{\textbf{+}}            & \multicolumn{1}{c|}{1.84}       & \multicolumn{1}{c|}{}                      & \multicolumn{1}{c|}{-0.15}      & \multicolumn{1}{c|}{\textbf{*}}            & \multicolumn{1}{c|}{-0.11}      & \multicolumn{1}{c|}{\textbf{+}}            & \multicolumn{1}{c|}{-0.11}      & \multicolumn{1}{c|}{\textbf{+}}            & \multicolumn{1}{c|}{0.04}      &                       \\
\multicolumn{1}{|c|}{12}               & \multicolumn{1}{c|}{2.06}       & \multicolumn{1}{c|}{\textbf{+}}            & \multicolumn{1}{c|}{3.12}       & \multicolumn{1}{c|}{\textbf{*}}            & \multicolumn{1}{c|}{-0.03}      & \multicolumn{1}{c|}{}                      & \multicolumn{1}{c|}{0.03}       & \multicolumn{1}{c|}{}                      & \multicolumn{1}{c|}{0.03}       & \multicolumn{1}{c|}{}                      & \multicolumn{1}{c|}{0.00}      &                       \\ \hline \\
\multicolumn{13}{c}{\textbf{(a) Columbia games corpus (CGC)}}        \\                                                                                                                                                                                                                                                                                                                                                                                                                                     \\ \hline
\multicolumn{1}{|c|}{1}                & \multicolumn{1}{c|}{-0.21}      & \multicolumn{1}{c|}{}                      & \multicolumn{1}{c|}{0.65}       & \multicolumn{1}{c|}{}                      & \multicolumn{1}{c|}{-0.09}      & \multicolumn{1}{c|}{\textbf{+}}            & \multicolumn{1}{c|}{0.06}       & \multicolumn{1}{c|}{}                      & \multicolumn{1}{c|}{-0.01}      & \multicolumn{1}{c|}{}                      & \multicolumn{1}{c|}{0.07}       &                       \\
\multicolumn{1}{|c|}{2}                & \multicolumn{1}{c|}{0.03}       & \multicolumn{1}{c|}{}                      & \multicolumn{1}{c|}{4.39}       & \multicolumn{1}{c|}{\textbf{*}}            & \multicolumn{1}{c|}{0.15}       & \multicolumn{1}{c|}{\textbf{*}}            & \multicolumn{1}{c|}{0.09}       & \multicolumn{1}{c|}{}                      & \multicolumn{1}{c|}{-0.18}      & \multicolumn{1}{c|}{\textbf{+}}            & \multicolumn{1}{c|}{0.05}       &                       \\
\multicolumn{1}{|c|}{3}                & \multicolumn{1}{c|}{1.33}       & \multicolumn{1}{c|}{}                      & \multicolumn{1}{c|}{3.32}       & \multicolumn{1}{c|}{\textbf{*}}            & \multicolumn{1}{c|}{0.03}       & \multicolumn{1}{c|}{}                      & \multicolumn{1}{c|}{0.02}       & \multicolumn{1}{c|}{}                      & \multicolumn{1}{c|}{-0.05}      & \multicolumn{1}{c|}{}                      & \multicolumn{1}{c|}{-0.04}      &                       \\
\multicolumn{1}{|c|}{4}                & \multicolumn{1}{c|}{-0.38}      & \multicolumn{1}{c|}{}                      & \multicolumn{1}{c|}{-0.04}      & \multicolumn{1}{c|}{}                      & \multicolumn{1}{c|}{0.09}       & \multicolumn{1}{c|}{\textbf{+}}            & \multicolumn{1}{c|}{0.01}       & \multicolumn{1}{c|}{}                      & \multicolumn{1}{c|}{-0.01}      & \multicolumn{1}{c|}{}                      & \multicolumn{1}{c|}{0.08}       &                       \\
\multicolumn{1}{|c|}{5}                & \multicolumn{1}{c|}{-0.02}      & \multicolumn{1}{c|}{}                      & \multicolumn{1}{c|}{0.94}       & \multicolumn{1}{c|}{}                      & \multicolumn{1}{c|}{-0.11}      & \multicolumn{1}{c|}{\textbf{+}}            & \multicolumn{1}{c|}{-0.06}      & \multicolumn{1}{c|}{}                      & \multicolumn{1}{c|}{-0.09}      & \multicolumn{1}{c|}{}                      & \multicolumn{1}{c|}{0.00}      &                       \\
\multicolumn{1}{|c|}{6}                & \multicolumn{1}{c|}{-3.88}      & \multicolumn{1}{c|}{\textbf{*}}            & \multicolumn{1}{c|}{-3.86}      & \multicolumn{1}{c|}{\textbf{*}}            & \multicolumn{1}{c|}{0.05}       & \multicolumn{1}{c|}{}                      & \multicolumn{1}{c|}{0.00}    & \multicolumn{1}{c|}{}                      & \multicolumn{1}{c|}{0.06}       & \multicolumn{1}{c|}{}                      & \multicolumn{1}{c|}{0.11}       &                       \\
\multicolumn{1}{|c|}{7}                & \multicolumn{1}{c|}{3.14}       & \multicolumn{1}{c|}{\textbf{*}}            & \multicolumn{1}{c|}{4.62}       & \multicolumn{1}{c|}{\textbf{*}}            & \multicolumn{1}{c|}{-0.09}      & \multicolumn{1}{c|}{}                      & \multicolumn{1}{c|}{-0.18}      & \multicolumn{1}{c|}{\textbf{*}}            & \multicolumn{1}{c|}{-0.07}      & \multicolumn{1}{c|}{}                      & \multicolumn{1}{c|}{0.11}       &                       \\
\multicolumn{1}{|c|}{8}                & \multicolumn{1}{c|}{1.86}       & \multicolumn{1}{c|}{}                      & \multicolumn{1}{c|}{3.86}       & \multicolumn{1}{c|}{\textbf{*}}            & \multicolumn{1}{c|}{0.10}       & \multicolumn{1}{c|}{\textbf{*}}            & \multicolumn{1}{c|}{-0.02}     & \multicolumn{1}{c|}{}                      & \multicolumn{1}{c|}{0.07}       & \multicolumn{1}{c|}{}                      & \multicolumn{1}{c|}{0.12}       & \textbf{*}            \\
\multicolumn{1}{|c|}{9}                & \multicolumn{1}{c|}{0.67}       & \multicolumn{1}{c|}{}                      & \multicolumn{1}{c|}{2.08}       & \multicolumn{1}{c|}{\textbf{+}}            & \multicolumn{1}{c|}{0.04}       & \multicolumn{1}{c|}{}                      & \multicolumn{1}{c|}{0.03}       & \multicolumn{1}{c|}{}                      & \multicolumn{1}{c|}{0.01}       & \multicolumn{1}{c|}{}                      & \multicolumn{1}{c|}{0.02}       &                       \\ \hline  \\
\multicolumn{13}{c}{\textbf{(b) SK Games corpus}}   \\             
\end{tabular}%
    }
    \caption{Summary of entrainment results on auditory and semantic entrainment in (a) Columbia games corpus and (b) Sk-games corpus with significant results after Bonferroni correction (*) with $\alpha = 0.004$ and $0.005$ for the English and Slovak and without  Bonferroni correction (+) with $\alpha = 0.05$ entrainment type (proximity, convergence, and synchrony)}
    \label{tab:CGC}
\end{table}
\subsubsection{Columbia-Games corpus}
Table \ref{tab:CGC} (a) shows the auditory and semantic entrainment results in the Columbia games corpus.

\textit{Proximity:} The English dataset shows little evidence of local proximity on auditory features. Only three sessions shows evidence of positive proximity. On the semantic level, in contrast, we found seven sessions that showed positive proximity. In addition, we observed that the distribution of the sessions with positive proximity in two levels is not random and that in all but one case, if people entrain on the auditory level they also entrain on the semantic level.

\textit{Convergence:} We found little evidence of convergence on both levels in the Columbia games corpus. In auditory features, only one session shows significant evidence of divergence, i.e., differences between partners increase over time. On the contrary, one session shows significant evidence of positive convergence in semantic features. 

\textit{Synchrony:} The auditory features showed little evidence of synchrony. Only one session shows evidence of positive synchrony. Positive synchrony implies both the speakers are moving in the same direction, i.e., if speaker A raises his/her voice, then speaker B also raises his/her voice. On the contrary, we did not find evidence of synchrony on semantic features in the English corpus. Furthermore, before the Bonferroni correction, we found that two sessions showed negative synchrony; one session exhibited positive synchrony in semantic features, and one session exhibited positive synchrony in auditory features.

\subsubsection{SK-Games Corpus}
Table \ref{tab:CGC} (b) shows the results of auditory and semantic entrainment with proximity, convergence, and synchrony as entrainment metrics based on the Slovak games corpus.

\textit{Proximity:} The Slovak data shows little evidence of proximity on the auditory level. For auditory features, only one session shows evidence of positive proximity, and only one shows negative proximity. On the contrary, four sessions show significant positive proximity for semantic features, while one shows significant negative proximity. In addition, we observed a similar pattern that we observed earlier in the English corpus, i.e., people entrain on both features when they entrain on auditory features.

\textit{Convergence:} We found little evidence of convergence on both levels in the Slovak data. Two sessions display evidence of positive convergence for auditory features. On the contrary, only one session showed evidence of divergence on the semantic level.
Additionally, before applying the Bonferroni correction, we found that people converge on auditory features more when compared to semantic features in the SK-games corpus.

\textit{Synchrony:} The Slovak data shows little evidence of synchrony on semantic features: One session shows positive synchrony for semantic features. On the contrary, no session shows evidence of synchrony in auditory features. In \cite{levitan2015a}, the authors reported negative synchrony is evident on almost every para-linguistic (auditory) feature of the SK-games corpus. We found similar evidence to be true where 6 out of 9 sessions show negative synchrony; however, they are not statistically significant.

\subsection{Relationship between auditory and semantic entrainment}

We measured two sets of adjacent distances using (Eq. \ref{equation:eq1}): a set of adjacent distances on auditory features and another set of adjacent distances on semantic features. We measured Pearson's correlation between adjacent distance on auditory and semantic embeddings to investigate the relationship between semantic and auditory features. 
\begin{table}[]
\centering
    \resizebox{8.25cm}{!}{
\begin{tabular}{cccccccccccc}
\cline{1-4} \cline{6-8} \cline{10-12}
\multicolumn{1}{c|}{Session} & \multicolumn{1}{c|}{r}    & \multicolumn{1}{c|}{p-value} & \multicolumn{1}{c}{Sig.} & \multicolumn{1}{c}{} & \multicolumn{1}{c|}{r}    & \multicolumn{1}{c|}{p-value} & \multicolumn{1}{c}{Sig.} & \multicolumn{1}{c}{} & \multicolumn{1}{c|}{r}    & \multicolumn{1}{c|}{p-value} & \multicolumn{1}{c}{Sig.} \\ \cline{1-4} \cline{6-8} \cline{10-12} 
\multicolumn{1}{c|}{1}        & \multicolumn{1}{c|}{0.02} & \multicolumn{1}{c|}{0.738}   &                           &                       & \multicolumn{1}{c|}{0.14} & \multicolumn{1}{c|}{0.038}   & +                         &                       & \multicolumn{1}{c|}{0.45} & \multicolumn{1}{c|}{0.000}   & *                         \\
\multicolumn{1}{c|}{2}        & \multicolumn{1}{c|}{0.13} & \multicolumn{1}{c|}{0.004}   & +                         &                       & \multicolumn{1}{c|}{0.16} & \multicolumn{1}{c|}{0.000}   & *                         &                       & \multicolumn{1}{c|}{0.44} & \multicolumn{1}{c|}{0.000}   & *                         \\
\multicolumn{1}{c|}{3}        & \multicolumn{1}{c|}{0.12} & \multicolumn{1}{c|}{0.008}   & +                         &                       & \multicolumn{1}{c|}{0.17} & \multicolumn{1}{c|}{0.000}   & *                         &                       & \multicolumn{1}{c|}{0.45} & \multicolumn{1}{c|}{0.000}   & *                         \\
\multicolumn{1}{c|}{4}        & \multicolumn{1}{c|}{0.17} & \multicolumn{1}{c|}{0.000}   & *                         &                       & \multicolumn{1}{c|}{0.19} & \multicolumn{1}{c|}{0.000}   & *                         &                       & \multicolumn{1}{c|}{0.36} & \multicolumn{1}{c|}{0.000}   & *                         \\
\multicolumn{1}{c|}{5}        & \multicolumn{1}{c|}{0.23} & \multicolumn{1}{c|}{0.001}   & *                         &                       & \multicolumn{1}{c|}{0.17} & \multicolumn{1}{c|}{0.012}   & +                         &                       & \multicolumn{1}{c|}{0.42} & \multicolumn{1}{c|}{0.000}   & *                         \\
\multicolumn{1}{c|}{6}        & \multicolumn{1}{c|}{0.17} & \multicolumn{1}{c|}{0.004}   & *                         &                       & \multicolumn{1}{c|}{0.22} & \multicolumn{1}{c|}{0.000}   & *                         &                       & \multicolumn{1}{c|}{0.50} & \multicolumn{1}{c|}{0.000}   & *                         \\
\multicolumn{1}{c|}{7}        & \multicolumn{1}{c|}{0.19} & \multicolumn{1}{c|}{0.000}   & *                         &                       & \multicolumn{1}{c|}{0.22} & \multicolumn{1}{c|}{0.000}   & *                         &                       & \multicolumn{1}{c|}{0.32} & \multicolumn{1}{c|}{0.000}   & *                         \\
\multicolumn{1}{c|}{8}        & \multicolumn{1}{c|}{0.12} & \multicolumn{1}{c|}{0.005}   & +                         &                       & \multicolumn{1}{c|}{0.15} & \multicolumn{1}{c|}{0.000}   & *                         &                       & \multicolumn{1}{c|}{0.30} & \multicolumn{1}{c|}{0.000}   & *                         \\  
\multicolumn{1}{c|}{9}        & \multicolumn{1}{c|}{0.11} & \multicolumn{1}{c|}{0.025}   & +                         &                       & \multicolumn{1}{c|}{0.15} & \multicolumn{1}{c|}{0.002}   & *                         & \multicolumn{1}{c}{} & \multicolumn{1}{c|}{0.23} & \multicolumn{1}{c|}{0.000}   & \multicolumn{1}{c}{*}    \\ \cline{10-12} 
\multicolumn{1}{c|}{10}       & \multicolumn{1}{c|}{0.30} & \multicolumn{1}{c|}{0.000}   & *                         &                       & \multicolumn{1}{c|}{0.25} & \multicolumn{1}{c|}{0.000}   & *                         &                       &                           &                              &                           \\
\multicolumn{1}{c|}{11}       & \multicolumn{1}{c|}{0.20} & \multicolumn{1}{c|}{0.000}   & *                         &                       & \multicolumn{1}{c|}{0.28} & \multicolumn{1}{c|}{0.000}   & *                         &                       &                           &                              &                           \\ 
\multicolumn{1}{c|}{12}      & \multicolumn{1}{c|}{0.11} & \multicolumn{1}{c|}{0.035}   & \multicolumn{1}{c}{+}    & \multicolumn{1}{c}{} & \multicolumn{1}{c|}{0.16} & \multicolumn{1}{c|}{0.002}   & \multicolumn{1}{c}{*}    &                       &                           &                              &                           \\ \cline{1-4} \cline{6-8}
                              &                           &                              &                           &                       &                           &                              &                           &                       &                           &                              &                           \\
\multicolumn{4}{c}{1) Using SBERT model}                                                                        &                       & \multicolumn{3}{c}{2) Using USE model}                                               & \multicolumn{3}{c}{}                                                             &                           \\
\multicolumn{1}{l}{}          & \multicolumn{1}{l}{}      & \multicolumn{1}{l}{}         & \multicolumn{1}{l}{}      & \multicolumn{1}{l}{}  & \multicolumn{1}{l}{}      & \multicolumn{1}{l}{}         & \multicolumn{1}{l}{}      & \multicolumn{1}{l}{}  & \multicolumn{1}{l}{}      & \multicolumn{1}{l}{}         & \multicolumn{1}{l}{}      \\
\multicolumn{8}{c}{a) Columbia games corpus}                                                                                                                                                                                        &                       & \multicolumn{3}{c}{b) SK-games corpus}                                                                    
\end{tabular} 
}
     \caption{Entrainment relationship between semantic and auditory features in Columbia Games Corpus and SK-games corpus after Bonferroni correction (*) with $\alpha = 0.004$ and $0.005$ for the English and Slovak and without Bonferroni correction (+) with $\alpha = 0.05$ }
     \label{tab:CGC1}
\end{table}

\textit{Columbia Games Corpus:} Table \ref{tab:CGC1} (left panel 1a) shows results for the entrainment relationship between auditory and semantic features using the SBERT model in English Data. We found six sessions out of 12 exhibits a slightly significant positive correlation (mean \textit{r}=0.21). To explore the potential effect of the selection of language models (semantic model), we also utilized Google's Universal sentence encoder (USE) model \cite{cer2018a} for extracting semantic features for each turn. Using the USE model, we measured adjacent distance on semantic features and measured Pearson's correlation between semantic and auditory features.  Table \ref{tab:CGC1} (middle panel 2a) shows the entrainment relationship between auditory and semantic features for the USE model.  We found ten sessions out of 12 exhibits a slightly positive correlation (mean \textit{r}=0.20).  

\textit{SK-Games Corpus:} Table \ref{tab:CGC1} (rightmost panel b) shows that Slovak data has a stronger positive correlation between entrainment in both linguistic levels than the English data where all the sessions are positively correlated with mean \textit{r} = 0.40.
\section{Discussion and conclusion} 
We analyzed semantic and auditory entrainment using three different entrainment metrics over a total of 21 sessions of collaborative dyadic interactions in two languages. We observed the following patterns that emerged from the analysis. 

Firstly, proximity is more prevalent than synchrony and convergence in both semantic and auditory entrainment. In both languages, positive proximity is evident in a greater number of dialogues compared to convergence and synchrony, indicating the tendency of people to get closer to each other in both semantic and auditory space at a given point in time. 

Secondly, we found that semantic proximity is more prevalent than auditory proximity. In both datasets, we observed that people entrain on semantic features more when compared to auditory features. In general, when the semantics of two interlocutors become more similar, the interlocutor can understand the content of the conversation more easily. One possible reason for such a result can be traced to the type of corpora utilized for entrainment analysis. We used task-oriented corpora, where the objective was to communicate about specific items in order to reach a joint goal. Semantic entrainment is crucial in task-oriented conversations like this since the task cannot be completed successfully without it. In contrast, auditory entrainment is optional and may be used to support semantic entrainment or indicate various aspects of the negotiation in terms of social relationship between the interlocutors. The findings of our study might vary from analyzing entrainment in real-life conversational corpora where semantic and auditory entrainment might weigh differently.

Thirdly, we noticed that semantic and auditory entrainment are positively correlated. A positive relationship between different linguistic levels can be conceptualized as people who entrain on one level are more likely to entrain on other levels. This finding is consistent with the Interactive Alignment Model proposed by \cite{pickering_garrod_2004}. This cognitive theory suggests that alignment at one level leads to alignment at other levels. Our findings suggest entrainment can be considered a single latent behavior or a collection of linked behaviors where people aligning on auditory features are more likely to align on semantic features. It is interesting to note the directionality in our findings: semantic entrainment implies auditory one whereas the reverse is not the case. The results of our study may also inform models dealing with the percolation of entrainment across linguistic levels.

Lastly, we noted that selecting a language model is crucial in identifying the relationship between different linguistic levels. We measured the relationship between auditory and semantic linguistic levels using two different language models for extracting semantic features in the English dataset. We found variance in results where utilizing the SBERT model reported six sessions are significantly positively correlated with mean \textit{r} of 0.20. In contrast, the USE model reported that ten sessions are significantly positively correlated with mean \textit{r} = 0.21. The average results of correlations are almost identical (\textit{r} = 0.2 and 0.21); however, the number of sessions that are significantly positively correlated is different. A language model might account for such variability in results and when considering the entire corpus, differences are smoothed out. 

In the Slovak dataset, we found a relatively stronger correlation between auditory and semantic entrainment with mean \textit{r} = 0.40 on all sessions. It remains to be explored if this difference stems from the difference among the patterns of entrainment in Slovak and English or if, in part, it might stem from the selection of the language model as both datasets in the current study are similar. We did not find any other language models trained in Slovak due lower NLP resources compared to English. Extracting semantic features from different language models could allow us to have a more meaningful comparison and understand if such a stronger correlation is due to the language model.

To conclude, in earlier studies researchers used fragmented features and different methods to measure entrainment, which might have contributed to the variation in results. We measured entrainment using the comparable methodology on different levels and in different languages, and our measures captured entrainment patterns that differ from previous studies, e.g.  \cite{liu2021a}. This further implies that methodology and features utilized for measuring entrainment play an important role in finding the relationship between different levels. In our future work, we plan to investigate entrainment relationships also on other linguistic levels, such as lexical and syntactic, and analyze the entrainment relationships among them. This will allow us to pursue developing SDS whose entrainment functionalities are informed by the relationship among entrainment on different linguistic levels, which could provide a more naturalistic conversational experience in future human-machine spoken interactions.

\section{Acknowledgements}
\ifinterspeechfinal
     This project has received funding from the European Union’s Horizon 2020 \textit{research and innovation programme under the Marie Skłodowska-Curie grant agreement No} 859588 and in part from the Slovak Granting Agency grant VEGA2/0165/21 and Slovak Research and Development Agency grant APVV-21-0373.
\else
     
\fi

\bibliographystyle{IEEEtran}
\bibliography{mybib}

\end{document}